\documentclass[11pt]{article}

\usepackage[final]{acl}

\usepackage{times}
\usepackage{latexsym}
\usepackage{amsmath,amssymb}      
\usepackage{booktabs}
\usepackage{algorithm}
\usepackage[noend]{algpseudocode}
\usepackage{enumitem}
\usepackage[most]{tcolorbox}
\usepackage{seqsplit}

\usepackage[T1]{fontenc}

\usepackage[utf8]{inputenc}

\usepackage{microtype}

\usepackage{inconsolata}

\usepackage{graphicx}

\title{NALA\_MAINZ at BLP-2025 Task 2: A Multi-agent Approach for Bangla Instruction to Python Code Generation}

\author{
  \textbf{Hossain Shaikh Saadi\textsuperscript{1}}\quad~ 
  \textbf{Faria Alam\textsuperscript{2}}\quad~ 
  \textbf{Mario Sanz-Guerrero\textsuperscript{1}}\\
  \textbf{Minh Duc Bui\textsuperscript{1}}\quad~ 
  \textbf{Manuel Mager\textsuperscript{1}}\quad~ 
  \textbf{Katharina von der Wense\textsuperscript{1,3}}\\
  \textsuperscript{1}Johannes Gutenberg University Mainz, Germany\\
  \textsuperscript{2}Saarland University, Germany\quad~ 
  \textsuperscript{3}University of Colorado Boulder, USA\\
  \texttt{\href{mailto:hsaadi@uni-mainz.de}{hsaadi@uni-mainz.de}}
}

\begin{document}
\maketitle
\begin{abstract}
This paper presents the winning system for the BLP-2025 Shared Task on Code Generation from Bangla Instructions, which consists of a multi-agent pipeline. First, a code-generation agent produces an initial solution from the input instruction. The candidate program is then executed against the provided unit tests (pytest-style, assert-based). Only the failing cases are forwarded to a debugger agent, which reruns the tests, extracts error traces, and, conditioning on the error messages, the current program, and the relevant test cases, generates a revised solution. Using this approach, our submission achieves first place in the shared task with a $Pass@1$ score of 95.4. We make our code publicly available.\footnote{\url{https://github.com/shaikhsaadi999/blp25_code_genneration}}
\end{abstract}

\section{Introduction}

In recent years, the quality of automatically generated code based on natural language instructions has increased rapidly, and a plethora of work exists in this domain \cite{qwen3technicalreport,hui2024qwen2,starcoder,codesurvey2}. However, most benchmarks and systems remain vastly English-centric \cite{codesurvey1}. This imbalance narrows access to program-synthesis tools for non-English speakers and limits our understanding of how linguistic factors -- such as morphology, script variation, and code-mixing -- impact the path from instructions to executable programs. Additionally, it makes it harder for developers with limited English proficiency to use these tools \cite{mconala}. Bangla -- spoken by over 270 million people worldwide -- is an example of an inadequately supported language in this area. Although NLP resources are expanding \cite{devdata, testdata}, there are still few high-quality evaluations and systems for Bangla instruction-to-code generation \cite{testdata}. 

This paper presents the winning system for the BLP-2025 Code Generation Shared Task \cite{task2}. Our system adopts a targeted two-agent pipeline. First, a code-generation agent produces an initial Python solution from the Bangla instruction. We then execute the generated code against a set of pytest-style unit tests from the provided dataset and an external dataset \cite{austin2021program} to obtain concrete failure signals. Rather than re-prompting every data sample, only the failing samples together with the error traces, the current code, and the corresponding unit tests are passed to a debugger agent that proposes minimal, correctness-oriented edits. By localizing debugging to the right spots, we concentrate our inference effort and avoid needless code changes. We evaluate multiple proprietary APIs for this multi-agent approach. Empirically, this approach delivers strong results on the shared task data. 

\begin{figure*}[t]
  \centering
  \includegraphics[width=\textwidth]{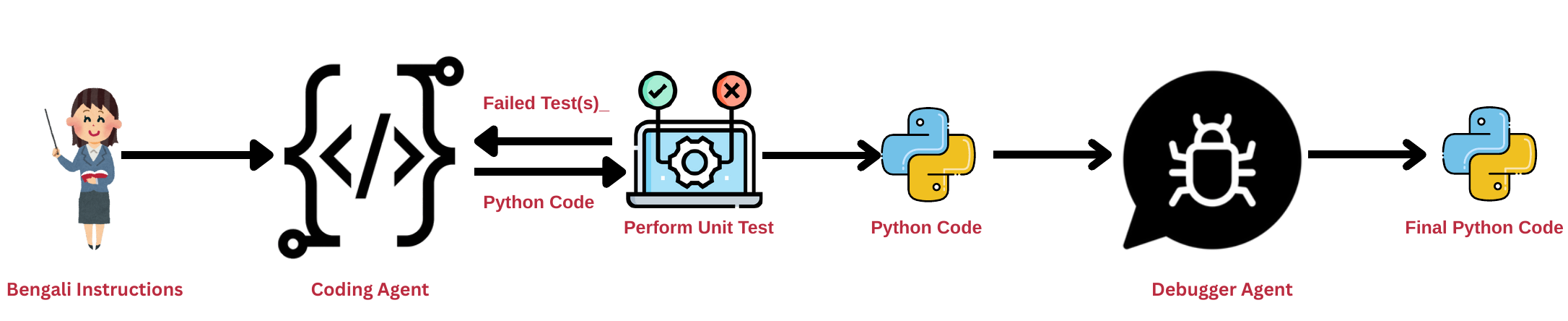}
  \caption{Multi-agent Bangla$\to$Python code generation pipeline with selective debugging via unit test feedback.}
  \label{fig:pipeline}
\end{figure*}

Our contributions are two-fold: (i) a simple but effective agent architecture that couples generation with test-driven refinement; (ii) a selective feedback mechanism that exposes the debugger only to failing cases and distilled execution traces.
On the official leaderboard, our system achieves 
the best performance among all participating teams (Pass@1: 95.4\%). We complement the results with a small ablation study and analysis. An overview of our proposed system is provided in Figure \ref{fig:pipeline}.

\section{Related Work}

Code generation with large language models (LLMs) is a well-established area of study~\cite{codesurvey1,codesurvey3,codegen1}, with interest and capability surging in recent years~\cite{qwen3technicalreport,hui2024qwen2,starcoder}. Despite this momentum, there remains relatively little work on Bangla instruction-to-code generation~\cite{testdata}. Test-driven development (TDD) is a well-established methodology in which developers write tests before implementing the functional code, ensuring that the resulting solution directly aligns with the initial problem statements~\cite{tdd}. In TDD, test cases are the central artifact, and there is a growing body of research on automated test-case generation~\cite{tddrl,hardtest,testeval,testcase1}. For evaluating code generation systems, unit test-based evaluation has become the de facto standard \cite{metric}. Rather than relying on text-only metrics used in machine translation (e.g., BLEU), running generated programs against hidden tests directly measures functional correctness~\cite{codesurvey2}.

\section{Task Definition}
The goal of the shared task is to generate Python code from a given Bangla instruction. The instruction itself contains the given Bangla instruction, function name, and argument names. Furthermore, assert-based pytest-style unit tests \cite{task2} are provided for each instruction. For each instance $i$, we are given a Bangla natural-language instruction $x_i \in X$ which includes
a function name and its argument names, and a \texttt{pytest} test suite $T_i$. No type signature is provided for the arguments of the given function; required behavior is determined solely by the instructions $x_i$ and test suites $T_i$. A candidate program must define the function precisely as provided by the function name since \texttt{pytest} is designed with the same function names. For example, if a function's name is \texttt{min\_cost}, for this function, one given assert-based unit test will be like this, \texttt{\seqsplit{assert min\_cost([[1,2,3],[4,8,2],[1,5,3]],2,2)==8}}. A program \emph{passes} if all tests in $T_i$ succeed, and the goal is to 
generate a program $\hat{p}_i$ for an instruction $x_i$ that passes all tests. For this task, the evaluation metric is $Pass@1$ \cite{metric}. This is a standard metric for code generation tasks that measures the proportion of instructions for which a single generated code passes all the provided unit tests. So, in short, given one attempt, how often does the model produce an entirely correct solution. In the most common setting, for a given instruction, one solution per problem is generated, and then the fraction of the generated solutions that pass all the unit tests is reported as a $Pass@1$ score.

\section{Data}\label{data}

\paragraph{Development dataset} Our development set consists of 400 Bangla instructions paired with function names, each accompanied by three unit tests. The organizers provide this dataset \cite{devdata}.

\paragraph{Test dataset} The test set contains 500 Bangla instructions with function names, each accompanied by a single unit test, so that we gain some understanding of the behavior of the input and output of the function we need to generate. 
Final submitted codes are tested on hidden unit tests for the sake of fair and robust evaluation.

\paragraph{External dataset} The task permits the use of external models and datasets. To augment the available tests for code generation, we search for a resource with matching function formats and existing unit tests. We identify \citet{austin2021program}, which directly covers unit tests for 480 relevant samples. In this external dataset, the number of unit tests per instance varies: some instances are provided with three tests, while others are provided with four. For each instruction in the provided data, we match its function name against those in \citet{austin2021program}, retrieve the corresponding 480 functions, and extract all associated unit tests. Any tests that do not overlap with the original test set are appended to the single provided unit test.

\section{Pipeline Overview} \label{pipeline}
We employ two agents: a \textbf{code generation agent} $G_\theta:\mathcal{X}\!\times\!\mathcal{T}\!\to\!\mathcal{P}$ that takes an instruction ${x}_i \in \mathcal{X}$ together with a test suite ${T}_i \in \mathcal{T}$ and returns a generated program. The provided test suite is not included in the prompt for the code generation agent; it is only used for testing. Function names and argument names are also provided inside the instruction. A \textbf{debugger agent} $D_\phi:\mathcal{X}\!\times\!\mathcal{T}\!\times \mathcal{P} \!\to\!\mathcal{P'}$ then takes an instruction ${x}_i \in \mathcal{X}$, a test suite ${T}_i \in \mathcal{T}$,
and a buggy program $p \in \mathcal{P}$ and returns an improved program.

\paragraph{Stage~1 (Code generation agent)} For each Bangla instruction $x_i$, this agent generates a Python code $p$ and executes each unit test from the provided \texttt{pytest} test suite $T_i$. If it fails for any of the tests, it generates again by mentioning in the prompt that it has failed. After this second step, the code generation agent saves the generated code for each instruction $x_i$. If a generated code fails even after the second attempt, it is saved as \textit{failed}. Otherwise, it is saved as \textit{passed}.

\paragraph{Stage~2 (Debugger agent)} In this stage, the debugger agent processes only those codes that fail in Stage 1. We call this set of failed codes $\mathcal{F}$. For each failed code $p\in\!\mathcal{F}$, we re-execute the provided test suites to consolidate the error trace. After that, the debugger conditions on its corresponding instruction $x_p$, test suite $T_p$, and the error traces $\widetilde{\mathcal{T}}(p)$ for the test suite to produce repaired code $p'$, which is saved as the final generated code. Our detailed algorithm is provided in Algorithm \ref{alg:two_step}.

\paragraph{Test case generation} Apart from these two agents, we employ another agent for test case generation from the given Bangla instruction $x_i$, and a \texttt{pytest} test suite $T_i$. For each instruction, we generate five extra test cases. We employ an abstract syntax tree-based syntax evaluation technique, which guarantees that the selected unit tests are syntactically correct. This agent is not part of our primary system.

\paragraph{Proprietary APIs} For our proposed pipeline, we use proprietary APIs from OpenAI, Google, and Anthropic. From Google, we use \texttt{Gemini-2.5-Flash}, from OpenAI, we use \texttt{GPT-5} and \texttt{GPT-4.1}, and from Anthropic, \texttt{Claude-Sonnet-4}. 
For our primary submission, we use \texttt{GPT-5} since it is the best-performing model in the dev set. For the code generation agent, we set the reasoning effort to \texttt{low} and for the debugger agent, we set it to \texttt{high}.

All detailed prompts for our agents are provided in the Appendix \ref{appendix}.

\begin{algorithm}[h]
\caption{Multi-agent Code Generation}
\label{alg:two_step}
\begin{algorithmic}[1]
\Require $\mathcal{X},\mathcal{T}$, $G_\theta$, $D_\phi$
\Ensure Final code $p'$

\Function{$\mathsf{RunTests}$}{$p,T$}
  \State Run \texttt{pytest} suite $T$ on $p$ (exact \texttt{name}/\texttt{args}).
  \State \Return (\textsc{Pass},$\varnothing$) if all pass; else (\textsc{Fail}, aggregated trace $\widetilde{\mathcal{T}}(p)$).
\EndFunction
\State \textbf{function end}
\For{$i=1$ \textbf{to} $N$} \\
\Comment{Stage 1: Code Generation (max 2 attempts)}
  \State ${x}_i \gets$ Instruction for instance $i$
  \State ${T}_i \gets$ Test suite for instance $i$
  \State $p \gets G_\theta(x_i)$;
  \State $(r,\_)\gets \mathsf{RunTests}(p,T_i)$
  \If{$r=\textsc{FAIL}$} \State $p \gets G_\theta(x_i)$;
  \State $(r,\_)\gets \mathsf{RunTests}(p,T_i)$
  \EndIf
  \State SAVE($p$)
\EndFor

\For{each $p\in\mathcal{F}$} \Comment{Stage 2: Debugger on failures only, here $\mathcal{F}$ is the set of all failed codes from the Stage 1}
  \State ${x}_p \gets$ Corresponding Instruction for $p$
  \State ${T}_p \gets$ Corresponding Test suite for $p$
  \State $(r,\widetilde{\mathcal{T}}(p))\gets \mathsf{RunTests}(p,T_p)$
  \If{$r=\textsc{FAIL}$}
  \State $p' \gets D_\phi(x_p,T_p,\widetilde{\mathcal{T}}(p))$
  \EndIf
  \State SAVE($p'$)
\EndFor
\end{algorithmic}
\end{algorithm}

\begin{table}[t]
\centering
\small
\begin{tabular}{lcc}
\toprule
\textbf{Model} & \textbf{Pass@1} & \textbf{Error rate} \\
\midrule
GPT-5            & \textbf{64.60} & \textbf{35.40} \\
Gemini-2.5 Flash   & 52.60         & 47.40 \\
Claude sonnet 4  & 58.20         & 41.80 \\
GPT-4.1          & 58.00          & 42.00 \\
\bottomrule
\end{tabular}
\caption{Stage-1 (code generation only; Pass@1 on Test Set). Error rate is $100-\text{Pass@1}$.}
\label{tab:stage1-main}
\end{table}

\begin{table}[t]
\centering
\small
\begin{tabular}{lcc}
\toprule
\textbf{Model} & \textbf{Pass@1} & \textbf{Error rate} \\
\midrule
GPT-5            & \textbf{95.4} & \textbf{4.6} \\
Gemini-2.5 Flash   & 59.80         & 40.20 \\
Claude sonnet 4  & 79.00         & 21.00 \\
GPT-4.1          & 82.60          & 17.40 \\
\bottomrule
\end{tabular}
\caption{Stage-2 (debugging with error traces and unit tests; Pass@1 on Test Set). Error rate is $100-\text{Pass@1}$.}
\label{tab:stage2-main}
\end{table}

\section{Results \& Discussion}
We separately report our scores for Stage 1 (coding agent) and Stage 2 (debugger agent) in Table \ref{tab:stage1-main} and Table \ref{tab:stage2-main}, respectively. We find that without the 
unit tests, any prompt we give to the model fails to provide the correct code almost 40\% of the time. Only \texttt{GPT-5} can go above 60\% in $Pass@1$ score; even \textsc{Claude Sonnet 4} can not go over 60\%. When the test cases are presented to the model within the prompt during the debugging process in the second stage, the model's performance improves significantly, except for \texttt{Gemini-2.5-Flash}. For this model, the $Pass@1$ score improves only 13.68\%, which is very low compared to the other three models.
\texttt{GPT-5} attains the top Stage-1 $Pass@1$ of \textbf{64.6}, outperforming \texttt{GPT-4.1} by 6.6 points, \texttt{Claude Sonnet 4} by 6.4, and \texttt{Gemini-2.5 Pro} by 12.0. Full Stage-1 results are shown in Table~\ref{tab:stage1-main}.

In stage two, we run the test cases first, then provide the debugger with the corresponding code and its related instructions, which fail at one of the unit tests. We give the error trace and all the unit tests. We observe that when unit tests are provided, the increase in $Pass@1$ scores is significantly higher than when the code is generated without the test cases in the prompt. We observe that for GPT-5, the increment is 47.67\% for \texttt{Gemini-2.5-Flash, Calude Sonnet 4, GPT-4.1}, 13.68\%, 35.73\%, and 41.37\%, respectively. So the biggest gain is from \texttt{GPT-5}. In stage 1, we only provide the Bengali instruction and the function name, which is not enough context for the model. That's why, when we add the unit test, the model's performance improves significantly.  
From this, it’s clear that more information about the problem context helps the model by providing the structure it needs to reason, cover edge cases, and self-debug. 

\paragraph{Overfitting to the unit tests} The scores indicate the effectiveness of incorporating the two-stage approach and augmenting test cases within the prompt. Also, there is a possibility that the models are being overfit to the provided test cases. During the development phase, we are provided with all the test cases that are used during the system's evaluation in Codabench.\footnote{\url{https://www.codabench.org/}} Our two-stage system achieve a 99.8 $Pass@1$ score during the evaluation of the development phase. We use all the provided unit tests in the prompt, which are used during the evaluation. However, during the test phase, we observe that 
when we submit our system for evaluation, there is a significant error rate compared to the development phase due to the presence of hidden unit tests. Using the available unit tests, we can achieve a 99.2 $Pass@1$ score using \textsc{GPT-5} and the evaluation script provided by the organizers locally; however, when we submit our system to Codabench to run on hidden unit tests, we obtain 95.4, which indicates that the generated codes sometimes can not handle edge cases, or they are not generalizing well, and are over-fitted to the provided unit tests. 

\paragraph{Effect of external data} In the provided test set, there is only one unit test for each instruction. We collect external test cases for 480 instructions out of 500 instructions, which we describe in Section~\ref{data}.\footnote{We used this external dataset after consulting with the organizers.} We observe that without the external unit test for \texttt{GPT-5}, our proposed pipeline achieves an 86.00 $Pass@1$ score, which is significantly lower than the 95.4 $Pass@1$ score achieved by augmenting unit tests from the external dataset. This indicates the effectiveness of the more unit tests, which helps the model to generate more generalized code. 

\paragraph{Effect of generated unit tests} As we mentioned in Section \ref{pipeline}, we also try generating test cases from the single provided test cases. We use the provided Bengali instruction, and a single test case to generate five test cases using \texttt{GPT-5} with a high reasoning effort. Out of 500 instructions, the model was able to generate at least 3 test cases for 461 instructions. Then we create the code in the same process. First, we generate without the test cases. Then we debug the code using the original and the augmented generated test cases. With the artificial test cases, we achieve an 84.00 $Pass@1$ score. Which is a lot higher than the stage 1 score (64.60) but lower than the stage 2 score (95.4) for \texttt{GPT-5}.
\begin{figure}[t]
    \centering
    \includegraphics[width=\columnwidth]{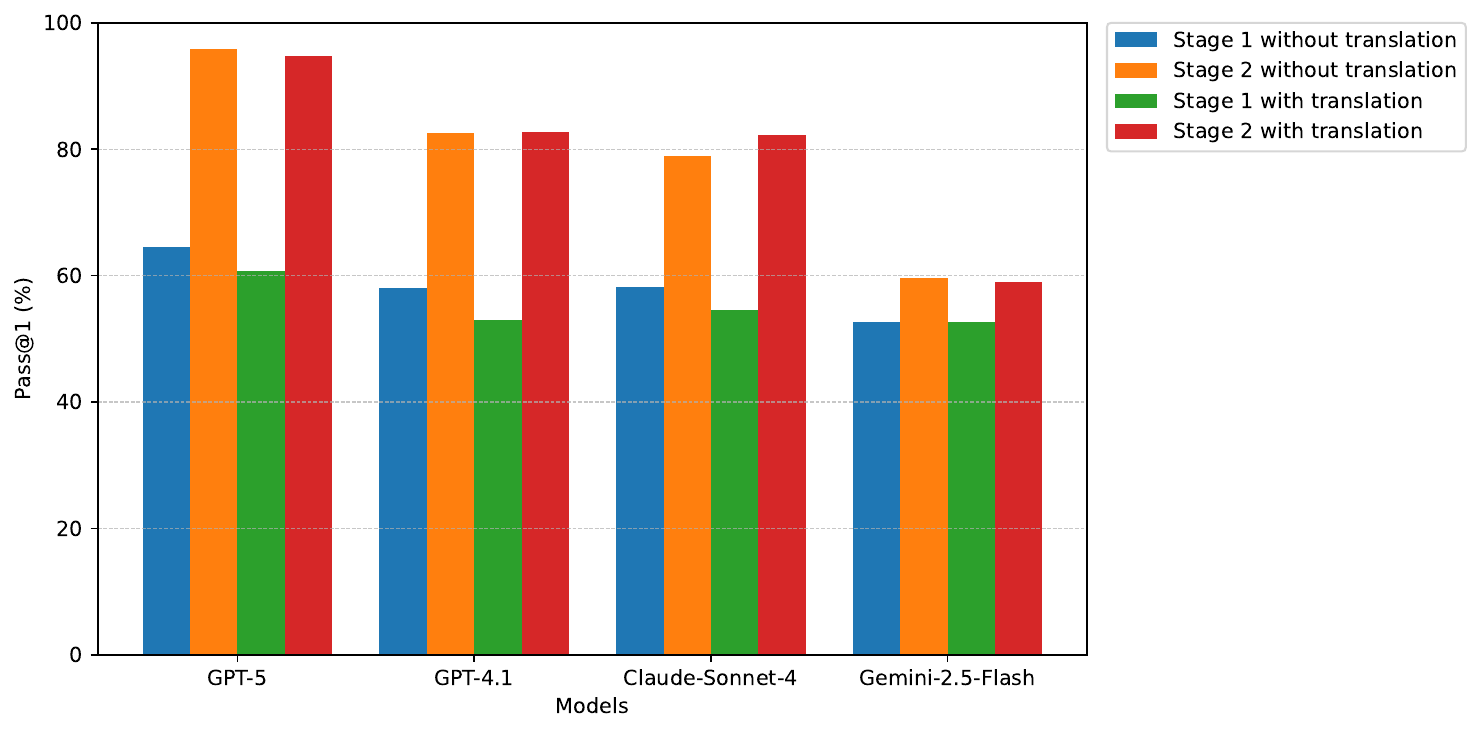}
    \caption{Pass@1 of different models across both stages, with and without translation.}
    \label{fig:pass1_models}
\end{figure}

\paragraph{Effect of translation} We translate the Bengali instructions into English using the \textsc{googletrans}\footnote{\url{https://pypi.org/project/googletrans/}} library, replacing the original Bengali instructions with the translated ones. Then we follow the same two-stage pipeline. In Figure~\ref{fig:pass1_models}, we report $Pass@1$ for four models (GPT-5, GPT-4.1, Claude-Sonnet-4, and Gemini-2.5-Flash) across two evaluation stages, with and without translation. Overall, GPT-5 achieves the highest $Pass@1$ in all conditions, followed by GPT-4.1, while Claude-Sonnet-4 and Gemini-2.5-Flash lag behind. For \texttt{Gemini-2.5-Flash}, we observe no change in either stage. In contrast, for \texttt{GPT-4.1}, the $ Pass@1$ score for the Stage-1 coding agent decreases from 58.00 to 53.00, while for Stage-2, it increases from 82.60 to 82.80. We see the same phenomenon for \texttt{Claude-Sonnet-4} as well. For Stage 1 $Pass@1$ decrease from 58.20 to 54.60 and for Stage 2 it increased from 79.00 to 82.20. However, for \texttt{GPT-5} in both stages, the translation decreases performance. The decrease is much higher for stage 1. In Stage 1 the $Pass@1$ decreases from 64.60 to 60.80, and for Stage 2, it decreases from 95.80 to 94.80. These results suggest that during translation, some task-specific information is lost, which in turn hinders the Stage-1 coding agent—but can also clarify particular instructions. When test cases are presented alongside the English instructions in Stage 2, this additional signal slightly improves scores despite the decline in Stage 1. Overall, different models react differently to translation, with varying sensitivity.

\section{Conclusion}
This paper presents our system for the BLP-25 Shared Task on Code Generation from Bangla Instructions: a straightforward and effective multi-agent pipeline. It couples an initial coder with a selective debugger driven by unit tests and their error traces. Based on \textsc{GPT-5}, our system achieves a $Pass@1$ score of 95.4\% and secures first place for the BLP-2025 Task 2. Our study highlights how structured feedback (error traces + tests) consistently boosts model performance over using only the instruction. Overall, the system demonstrates that targeted, test-driven refinement substantially improves code synthesis in an underserved language.

\section*{Acknowledgement}
This work was supported by the Carl Zeiss Foundation through the TOPML and MAINCE projects
(grant numbers P2021-02-014 and P2022-08-009I). We thank the anonymous reviewers for their feedback and suggestions.

\section*{Limitations}
Our study focuses exclusively on proprietary models for this task and does not evaluate any open-source models, which limits the reproducibility and generalizability of our findings. In addition, to enriching the number of unit tests supporting our test case-driven solution, we rely on an external dataset. Without this enrichment, models achieve approximately 10 percentage points lower scores than the reported ones. Finally, we observe substantial differences in performance across the proprietary models; however, we lack transparency into their training data and procedures, making it difficult to attribute these differences to specific factors. We utilize AI assistants, specifically ChatGPT (GPT-5), to assist with editing sentences in our paper writing.

\bibliography{custom}
\clearpage
\appendix

\section{Prompts}\label{appendix}


\begin{tcblisting}{
  enhanced,
  breakable,
  listing only,
  title=Coding Agent User Prompt,
  colback=white,
  colframe=blue!60!black,
  colbacktitle=blue!12,
  coltitle=blue!40!black,
  fonttitle=\bfseries\sffamily,
  boxrule=0.8pt,
  arc=3mm,
  width=\linewidth,
  center title=false,
  left=2mm, right=2mm, top=1mm, bottom=1mm, boxsep=1mm,
  listing options={
    basicstyle=\ttfamily\small,
    breaklines=true,
    breakatwhitespace=false,
    breakautoindent=false,
    breakindent=0pt,
    columns=fullflexible,
    keepspaces=true,
    showstringspaces=false,
    tabsize=2
  }
}
You are a strict Python coding assistant.
Write ONLY a valid Python implementation for the function below. Do not include tests, imports that are not used, main guards, comments, or explanations. Only provide the function definition and helper code inside the same file as needed.

{status}

Constraints:
Use the exact function name and signature.
Prefer clear, deterministic algorithms. No randomization.
Avoid printing and input() usage.
You may define additional helper functions and classes inside the same file if needed.
Please do not start the code with ```python

Function signature:
def {spec.name}({', '.join(spec.args)}):

Also consider the problem in Bengali:
{spec.instruction_bn}

Return ONLY the code block itself, with no backticks, quotes, or additional text.
\end{tcblisting}

\begin{tcblisting}{
enhanced,
breakable,
listing only,
title=Debugger Agent System Prompt,
colback=white,
colframe=blue!60!black,
colbacktitle=blue!12,
coltitle=blue!40!black,
fonttitle=\bfseries\sffamily,
boxrule=0.8pt,
arc=3mm,
center title=false,
halign=flush left,
left=0pt, right=0pt, top=1mm, bottom=1mm, boxsep=2pt,
width=\linewidth,
listing options={
basicstyle=\ttfamily\small,
breaklines=true,
breakatwhitespace=true,
breakautoindent=false,
breakindent=0pt,
xleftmargin=0pt
}
}
You are a Python debugging agent.
You receive the Instruction , current code, failing test(s), and the error trace.
Return ONLY corrected code. Keep changes minimal, focused to fix failures and consider edge cases.
Do not add comments or other non-code text.
You may define additional helper functions, classes and imports inside the same file if needed.
Please do not start the code with ```python
\end{tcblisting}

\begin{tcblisting}{
enhanced,
breakable,
listing only,
title=Debugger Agent User Prompt,
colback=white,
colframe=blue!60!black,
colbacktitle=blue!12,
coltitle=blue!40!black,
fonttitle=\bfseries\sffamily,
boxrule=0.8pt,
arc=3mm,
center title=false,
halign=flush left,
left=0pt, right=0pt, top=1mm, bottom=1mm, boxsep=2pt,
width=\linewidth,
listing options={
basicstyle=\ttfamily\small,
breaklines=true,
breakatwhitespace=true,
breakautoindent=false,
breakindent=0pt,
xleftmargin=0pt
}
}
Instruction (original): {instruction}
Current code: {code}
Failing tests: {failing_tests}
Error trace: {error_text}

Provide the fixed code only.
\end{tcblisting}

\newpage
\begin{tcblisting}{
enhanced,
breakable,
listing only,
title=Unit Test Generation Agent System Prompt,
colback=white,
colframe=blue!60!black,
colbacktitle=blue!12,
coltitle=blue!40!black,
fonttitle=\bfseries\sffamily,
boxrule=0.8pt,
arc=3mm,
center title=false,
halign=flush left,
left=1mm, right=1mm, top=1mm, bottom=1mm,
listing options={
basicstyle=\ttfamily\small,
breaklines=true,
breakatwhitespace=true,
breakautoindent=false,
breakindent=0pt,
xleftmargin=0pt
}
}
You are a Python unit test generation agent.
Return ONLY a Python list of strings, each a single pytest-style assert.
Think deeply and carefully about behavior and edge cases.
Generate constructive tests that reveal bugs, not trivial variations.
No explanations or code fences. Ensure asserts are syntactically valid.
\end{tcblisting}

\begin{tcblisting}{
enhanced,
breakable,
listing only,
title=Unit Test Generation Agent User Prompt,
colback=white,
colframe=blue!60!black,
colbacktitle=blue!12,
coltitle=blue!40!black,
fonttitle=\bfseries\sffamily,
boxrule=0.8pt,
arc=3mm,
center title=false,
halign=flush left,
left=1mm, right=1mm, top=1mm, bottom=1mm,
listing options={
basicstyle=\ttfamily\small,
breaklines=true,
breakatwhitespace=true,
breakautoindent=false,
breakindent=0pt,
xleftmargin=0pt
}
}
Target function name: {func_name}
Sample assert: {sample_assert}

Generate {num_tests} new, distinct, constructive unit tests (assert ...).
Use only Python literals in arguments; no I/O or imports.
\end{tcblisting}

\end{document}